\documentclass[11pt]{article}

\usepackage[preprint]{acl}

\usepackage{times}
\usepackage{latexsym}
\usepackage{multirow}
\usepackage{mathtools}
\usepackage{algorithm}
\usepackage{algorithmic}
\usepackage[T1]{fontenc}

\usepackage[utf8]{inputenc}

\usepackage{microtype}

\usepackage{inconsolata}

\usepackage{graphicx}

\title{Towards Principled Design of Mixture-of-Experts Language Models under Memory and Inference Constraints}

\author{Seng Pei Liew \and Kenta Shinzato \and Yuyang Dong \\
         SB Intuitions, Tokyo, Japan}

\begin{document}
\maketitle
\begin{abstract}
Modern Mixture-of-Experts (MoE) language models are designed based on total parameters (memory footprint) and active parameters (inference cost).
However, we find these two factors alone are insufficient to describe an optimal architecture.
Through a systematic study, we demonstrate that MoE performance is primarily determined by total parameters ($N_{total}$) and expert sparsity ($s:=n_{exp}/n_{topk}$).
 Moreover, $n_{exp}$ and $n_{topk}$ do not "cancel out" within the sparsity ratio; instead, a larger total number of experts slightly penalizes performance by forcing a reduction in core model dimensions (depth and width) to meet memory constraints.
This motivates a simple principle for MoE design which maximizes $N_{total}$ while minimizing $s$ (maximizing $n_{topk}$) and $n_{exp}$ under the given constraints.
Our findings provide a robust framework for resolving architectural ambiguity and guiding MoE design.
\end{abstract}

\section{Introduction}
\label{sec:intro}
Many publicly available Mixture-of-Experts (MoE) language models are characterized by total and active parameters, e.g., Qwen3-235B-A22B \cite{yang2025qwen3} and OLMoE-1B-7B \cite{muennighoff2024olmoe}.
This is likely due to the following practical deployment considerations:
Total parameters determine the model's memory footprint, while active parameters influence the computational cost during inference; both factors also fundamentally influence training latency.

Despite their prevalence, these two parameters are insufficient to fully characterize an MoE architecture.
To resolve this ambiguity, we define an MoE architecture using five variables: Number of layers or depth ($l$), hidden dimension or width ($d$), number of experts ($n_{ exp}$), number of activated experts ($n_{ topk}$), and the ratio of hidden dimension to expert hidden dimension (granularity, $g \coloneq d/d_{exp}$).
For a multi-head attention-based MoE model, the non-embedding parameter counts are: \footnote{For convenience, total and active parameters are always referred to as non-embedding parameters in this paper; a more precise description of configurations is given in Appendix \ref{app:exp_setup}.}
\begin{eqnarray}
    N_{total} &\approx ld^2\left(4 + 3  n_{exp}/g \right)  \label{eq:total}\\
    N_{active} &\approx ld^2\left(4 + 3  n_{topk}/g \right)  \label{eq:active}
\end{eqnarray}
Even if we constrain $N_{total}$ and $N_{active}$, there are still multiple choices of width, depth, and MoE configuration that satisfy the above two constraints.
This raises an important question: \emph{Given upper bounds on memory and inference, how should one select the variables to optimize performance?}

In this paper, we conduct a systematic study to identify the optimal configuration within a fixed budget of $N_{total}$ and $N_{active}$ ($N_{total} > N_{active}$).
 We show that as long as the width-to-depth ratio and granularity are within reasonable ranges, performance is mainly determined by $N_{total}$ and expert sparsity $s\coloneq n_{exp}/n_{topk}$.
 Furthermore, we identify a specific $n_{exp}$ penalty.
  For a fixed sparsity ratio, increasing $n_{exp}$ requires a compensatory reduction in the model's core dimensions ($l$ and $d$) to stay within the total parameter budget.
   This reduction leads to a net decrease in the active compute capacity per token, which slightly penalizes performance. 
These findings allow us to derive a principle that determines the optimal MoE configuration that maximizes $N_{total}$ while minimizing $s$ (maximizing $n_{topk}$) and $n_{exp}$ under the given constraints.
By fitting scaling laws across models and dataset sizes, we also test and validate our principle on two representative architectures.


\paragraph{Related work.}
 Various aspects of MoE models \cite{shazeer2017outrageously,lepikhin2020gshard,fedus2022switch} have been studied in the literature. 
 For a comprehensive survey of MoE models, please refer to \citet{cai2024survey}.
 We here only review the most relevant ones, which often relate the loss/performance of MoE models to the model configuration via scaling laws.

 \citet{clark2022unified} are among the first to propose a scaling law for MoE models, showing that the loss is a function of $N_{total}$ and $n_{exp}$.
 \citet{krajewski2024scaling,pmlr-v267-liew25a,abnar2025parameters,ludziejewski2025joint,tian2025towards} study aspects such as granularity, upcycling, sparsity, memory efficiency and dense-to-MoE efficiency ratio. 
While conclusions differ among studies, the consensus is that total parameters and/or active parameters are the most important factors affecting the performance of MoE models.
There is however no work trying to disentangle the interplay of all the parameters appearing in Equations \ref{eq:total} and \ref{eq:active} given total parameters and active parameters, as far as we know.

\paragraph{Setup.}
We use an architecture closely resembling that of Qwen3 \cite{yang2025qwen3}, which in turn follows the architecture of \cite{fedus2022switch}, scaling the feed-forward (FFN) layers of transformer-based language models into $n_{exp}$ experts and activating only $n_{topk}$ of them for each input token.
Furthermore, the fine-grained MoE architecture is adopted with the FFN hidden dimension $g$ times smaller than the attention hidden dimension \cite{dai2024deepseekmoe,krajewski2024scaling}.

The models are trained using a standard language modeling objective on the FineWeb-Edu dataset \cite{penedo2024fineweb}, with its performance evaluated on a held-out dataset.
More details about training hyperparameters and model configurations can be found in Appendix \ref{app:exp_setup}.

\section{Ablation fixing Total and Active Parameters}
As discussed above, total and active parameters are vital at specifying MoE models.
We first conduct ablation studies by fixing (approximately) these two parameters and varying other configurations to understand how different combinations of parameters change the loss.
\subsection{Granularity}
We study the effect of granularity $g$ by varying $n_{exp},n_{topk}$ in conjuction with $d_{exp}$ such that the total and active parameters are fixed.
Higher $g$ means finer granularity; i.e., smaller expert hidden dimension $d_{exp}$, but more experts.

Table~\ref{tab:ablate_g} shows the results of this ablation study on two different model sizes.
We see that $g$ of 4 to 8 generally minimize the loss, while increasing $g$ leads to diminishing returns.
Our results largely agree with \citet{tian2025towards} that $g$ (in our notations) between 4 and 6 is a good choice.
We use $g=4$ in the following experiments.

\begin{table}
  \centering
\begin{tabular}{cccccc}
\hline
$l$ & $d$ &  $g$ & $n_{exp}$ & $n_{topk}$ & Loss diff (\%) \\ \hline
\multirow{5}{*}{8} & \multirow{5}{*}{384}
   & 2 & 64 & 4 & 0.7\% \\
 &  & 4 & 128 & 8 & 0.48\% \\
 &  & \textbf{8} & \textbf{256} & \textbf{16} & \textbf{0.00\%} \\
 &  & 16 & 512 & 32 & 0.21\% \\ \hline
\multirow{5}{*}{18} & \multirow{5}{*}{1024}
  & 2 & 64 & 4 & 1.04\% \\
 &  & \textbf{4} & \textbf{128} & \textbf{8} & \textbf{0.00\%} \\
 &  & 8 & 256 & 16 & 0.29\% \\
 &  & 16 & 512 & 32 & 0.58\% \\ \hline
\end{tabular}
\caption{Ablation study on granularity $g$ with fixed total and active parameters. See Appendix \ref{app:ablation_details} for more results.}
\label{tab:ablate_g}
\end{table}

\subsection{Width-to-depth Ratio}
Next, we wish to study the choice of width-to-depth ratio $\gamma\coloneq d/l$.
We vary all relevant parameters in Equations \ref{eq:total} and \ref{eq:active} while keeping total and active parameters fixed approximately ($g$ is fixed to 4).

We see in Table \ref{tab:ablate_widthdepthratio} that models with the lowest $\gamma$ (e.g., 10) generally have higher loss, while those with mid-range values (e.g., 42) achieve the lowest loss in configurations explored. 
However, simply increasing $\gamma$ to the maximum (136) does not guarantee the lowest loss, suggesting a sweet spot.
We consequently choose to follow the standard setting of \citet{kaplan2020scaling} by keeping $\gamma$ between 32 and 64.

\begin{table}
  \centering
 \begin{tabular}{rrrrrr}
    \hline
$l$ & $d$ & $d/l$ & $n_{nexp}$ & $n_{topk}$ &  Loss diff (\%) \\
    \hline
16 & 240 & 15  & 43 & 4 & 0.65\%  \\
\textbf{8}  & \textbf{336} & \textbf{42}  & \textbf{43} & \textbf{4} & \textbf{0.00\%}  \\
4  & 480 & 120 & 43 & 4 & 1.12\%  \\
16 & 160 & 10  & 103 & 16 & 2.68\% \\
8  & 224 & 28  & 103 & 16 & 2.20\%  \\
4  & 320 & 80  & 103 & 16 & 1.68\%  \\
    \hline
\end{tabular}
  \caption{Ablation study on the width-to-depth ratio $\gamma=d/l$ with approximately the same total and active parameters. See Appendix \ref{app:ablation_details} for more results.}
  \label{tab:ablate_widthdepthratio}
\end{table}

\section{Scaling Laws for MoE Design}
\label{sec:main_results}
To systematically identify the optimal MoE architecture, we must navigate the five degrees of freedom ($l, d, n_{exp}, n_{topk}, g$) present in Equations \ref{eq:total} and \ref{eq:active}.
In the previous Section, we have constrained two of these dimensions to values that minimize loss while maintaining fixed total and active parameter budgets, reducing the remaining degrees of freedom to three; there is still ambiguity in determining the configuration.



In this Section, we vary the variables (under aforementioned constraints) across a wide range of model scales (30M to 3B parameters; details in Appendix \ref{app:scaling_law_details}), under a fixed training token budget, to study the loss behavior with respect to these variables, such that we can find the remaining (optimal) configuration that minimizes the loss.

 We assume a power-law relationship between loss and the variables \cite{hestness2017deep,hestness2019beyond}, such that we can perform linear regression on the log-log scale (e.g., $\log L = \alpha \log N_{total} + \beta$).
This allows us to perform statistically robust fitting and hypothesis testing on each variable's significance.

\subsection{Results}
\begin{figure}[t]
  \includegraphics[width=\columnwidth]{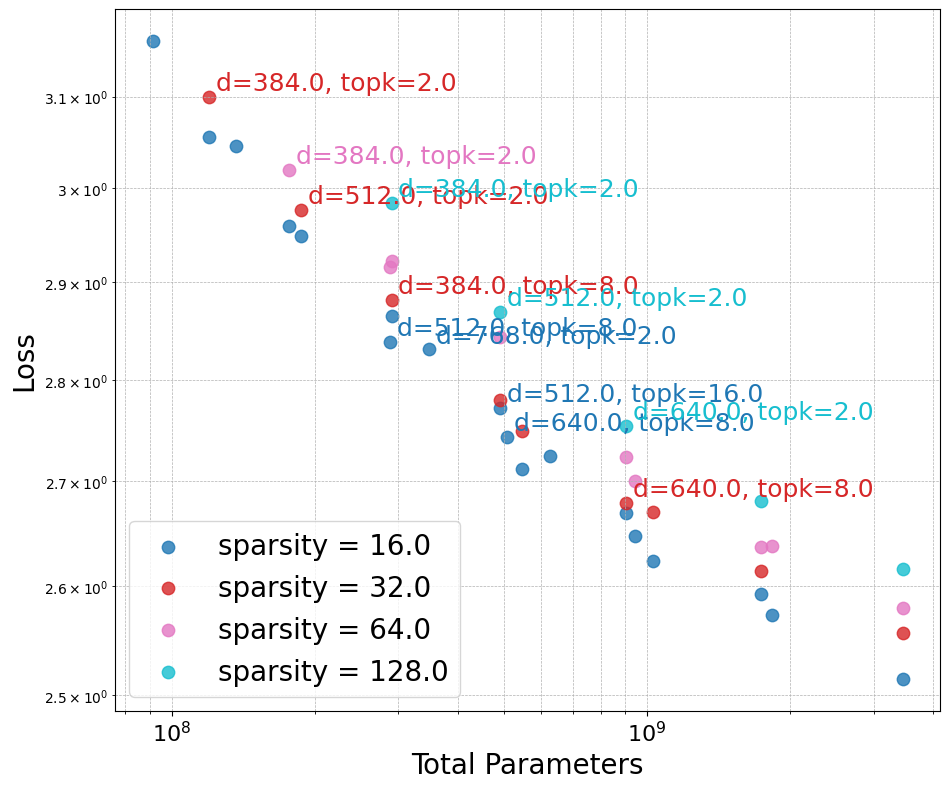}
  \caption{Loss vs. Total Parameters. Each point represents a model with a specific configuration.}
  \label{fig:loss_vs_ntotal}
\end{figure}

The loss values with respect to $N_{total}$ are shown in Figure \ref{fig:loss_vs_ntotal}.
We see that loss generally decreases with $N_{total}$ as a power law.
However, there is a large variance in loss values for similar $N_{total}$, suggesting that other factors are at play.
Moreover, optimizing loss based on $N_{total}$ alone cannot resolve for the optimal configuration.

We list the results of fitting various combinations of variables in Table \ref{tab:fitting_results}.
The performance of each function is mainly evaluated with the $R^2$ value of the fit, with additional scrutiny provided by t-tests for variable significance and condition numbers to detect collinearity.

Our analysis yields the following observations:
\begin{itemize}
  \item \textbf{Dominance of total parameters.} $N_{total}$ is a significantly stronger predictor of performance ($R^{2}=0.926$) than $N_{active}$ ($R^{2}=0.641$) when used as a single baseline. 
     \item \textbf{Total and active parameters.} While the interaction function $log(N_{total}) + log(N_{active}) + log(N_{total})log(N_{active})$ achieves the highest $R^{2}$ (0.988), it suffers from strong multicollinearity problem. 
  \item \textbf{Sparsity's role.} Adding $s$ to either $N_{total}$ or $N_{active}$ significantly improves the fit, achieving $R^2$ values of 0.983 and 0.944 respectively.
  Still, adding $s$ cannot resolve the ambiguity in MoE configuration because multiple combinations of ($n_{exp}, n_{topk})$ are still possible.
  \item \textbf{Disambiguated function.} We propose the use of $(N_{total}, n_{exp}, n_{topk})$ as the primary scaling law. This model achieves an $R^{2}$ of 0.985 with all variables being statistically significant and it fully resolves the MoE configuration.
\end{itemize}

The resulting fitted scaling law is as follows:
\begin{eqnarray}
    L &\propto {N_{total}}^{-0.052} {n_{exp}}^{0.023} {n_{topk}}^{-0.018} \\
    &= {N_{total}}^{-0.052} s^{0.018} {n_{exp}}^{0.005} \label{eq:scaling}
\end{eqnarray}

The fact that lower sparsity $s$ leads to better performance aligns with the intuition that activating more experts allows for better utilization of model capacity.
Furthermore, the $n_{exp}$ penalty can be interpreted as follows: To maintain a fixed $N_{total}$, the core dimensions ($l, d$) must be reduced to compensate for the larger $n_{exp}$.
 Because $N_{active}$ is scaled by these same dimensions, the net result is a decrease in active parameters per token, leading to a performance penalty due to reduced compute capacity.
 See Appendix \ref{app:math} for a more precise mathematical justification.

Our results imply the following principle for optimal configuration: Maximize $N_{total}$, $n_{topk}$ and minimize $n_{exp}$ to minimize the loss according to Equation \ref{eq:scaling} while satisfying constraints on $g$, $\gamma$, and $N_{active}$, leaving no ambiguity in the configuration.

Note however practical design is further constrained by the fact that the parameters in Equations \ref{eq:total} and \ref{eq:active} must be integers.
Additionally, $d$ must satisfy divisibility constraints due to attention head partitioning and parallelism (when tensor parallelism is used) requirements.
  Consequently, we propose the following iterative routine: First, we maximize $N_{total}$ by solving for the largest $(l,d)$ within memory, $g$ and partitioning constraints, looping over $\gamma,n_{exp}$.
   Second, we maximize $n_{topk}$ until the inference budget is saturated. 
   This routine ensures maximum leverage of available memory while preventing performance penalty incurred by excessive $n_{exp}$.
  We give a more detailed discussion in Appendix \ref{app:opt}.

\begin{table*}[ht]
\centering
\begin{tabular}{@{}lll@{}}
\hline
\textbf{Variables} & \textbf{$R^2$} & \textbf{Result/Intrepretation} \\ \hline
\textit{Simple Baseline} & & \\
$\log(N_{total})$ only & 0.926 & Baseline (Total params). \\
$\log(N_{active})$ only & 0.641 & Baseline (Active params). \\ \hline
\textit{Core Functions} & & \\
$\log(N_{active}) + \log(s)$ & 0.944 & Moderately good fit. \\
$\log(N_{total}) + \log(s)$ & 0.983 & Good fit. \\
$\log(N_{total}) + \log(n_{{exp}}) + \log(n_{topk})$ & \textbf{0.985} & \textbf{Good and disambiguated fit.} \\
$\log(N_{active}) + \log(n_{{exp}}) + \log(n_{topk})$ & 0.981 & Moderately good fit. \\ \hline
\textit{Interaction Functions} & & \\
$\log(N_{total}) + \log(s) + \log(N_{total})\log(s)$ & 0.983 & Interaction term redundant. \\
$\log(N_{total}) + \log(N_{active}) + \log(N_{total})\log(N_{active})$ & 0.988 & Strong multicollinearity problem. \\ \hline

\textit{Other Combinations} & & \\
$\log(N_{total}) + \log(N_{active}) + \log(s)$ & 0.985 &  $N_{active}$'s significance is low. \\
\end{tabular}
\caption{Results of fitting various combinations of variables to loss. More results in Table \ref{tab:more_fitting_results} of Appendix \ref{app:scaling_law_details}.}
  \label{tab:fitting_results}
\end{table*}

\subsection{Testing the Design Principle by comparing two Configurations}
The scaling law in Equation \ref{eq:scaling} is based on models trained with a fixed token budget.
To test its effectiveness in guiding MoE design, we conduct experiments by training models under various model and dataset sizes, fixing two configurations due to computational resource constraints.
Specifically, we compare MoE configurations with the same $s$: $(n_{exp},n_{topk})= (128, 8), (256, 16)$.
We fit Chinchilla-style scaling laws \cite{hoffmann2022training} for these configurations: 
\begin{eqnarray}
    L_{\text{128/8 or 256/16}} = A N_{total}^{-\alpha} + B D^{-\beta} + E \label{eq:chinchilla}
\end{eqnarray}
where $D$ is the dataset size (in tokens) and $A,B,E,\alpha,\beta$ are fitted coefficients.
According to Equation \ref{eq:scaling}, the $(128,8)$ configuration should outperform the $(256,16)$ one, when compared under the same $N_{total}$, $s$, and $D$, as it has smaller $n_{exp}$.

\begin{figure}
  \includegraphics[width=\columnwidth]{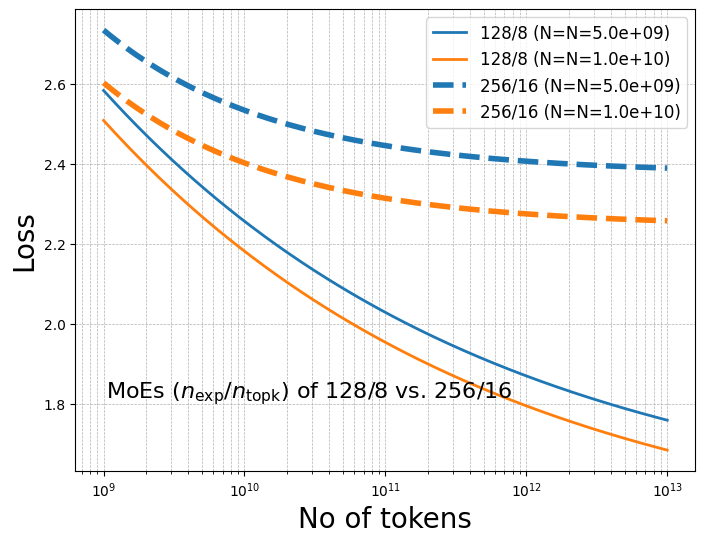}
  \caption{Comparison of loss between the $(128,8)$ and $(256,16)$ configurations with fitted scaling laws.}
  \label{fig:128_8_vs_256_16}
\end{figure}

We perform experiments and fit the scaling laws accordingly (see Appendix \ref{app:scaling_law_details} for details).
The resulting curves are shown in Figure \ref{fig:128_8_vs_256_16}.
Indeed, the $(128,8)$ configuration consistently outperforms the $(256,16)$ one under the same $N_{total}$ and $D$, confirming the effectiveness of our design principle.

We note that previous works \cite{abnar2025parameters,tian2025towards} have attempted to fit scaling laws involving model configuration and dataset size. 
However, they studied only $s$ without disentangling $n_{exp}$ and $n_{topk}$.
Our results indicate that a more complete scaling law involving all these factors would be desirable.

\section{Conclusion}
MoE models are mainstream for large language models \cite{jiang2024mixtral,dai2024deepseekmoe,yang2025qwen3}, yet their optimal configuration has remained poorly understood.
We disentangle the MoE configuration to show that $N_{total}$ is the primary predictor of performance, followed by expert sparsity $s$ and total experts $n_{exp}$.
Our design principle of optimizing these quantities under other conditions provides a rigorous foundation for building more efficient MoE models under practical constraints.

\section*{Limitations}
The first significant limitation is the scale of our experiments.
Due to limited computational resources, we are only able to conduct experiments on small to medium-sized models and datasets, with limited architecture search (we perform a kind of greedy search by first determining the optimal values of $g$ and $\gamma$ before moving on to deriving the scaling law) and hyperparameter tuning.
While we believe that our findings provide valuable insights into the design of MoE models, further validation on a larger scale, and more refined search/tuning is necessary to confirm their generalizability.
Ultimately, a full scaling law involving model configuration and dataset size would be desirable (which would require full grid search over model configuration, model size and dataset size).

The second limitation pertains to our use of total parameters and active parameters as proxies for memory and inference/training costs, respectively.
While these proxies are commonly used in the literature, they may not fully capture the complexities of real-world deployment scenarios.
Factors such as hardware architecture, software optimizations, parallelism training techniques, and specific use cases can significantly influence the actual memory footprint and inference latency of MoE models.

\section*{Acknowledgements}
We would like to thank Takuya Kato for helpful discussions and feedback.

\bibliography{custom}

\appendix
\section{More on Our Setup}
\label{app:exp_setup}
\subsection{Model Setup}
As described in the main text, we use a Qwen3-like MoE architecture \cite{yang2025qwen3}, adopting advances such as QK-norm \cite{henry2020query}, Rope positional encoding \cite{su2024roformer} beyond the original transformer architecture \cite{vaswani2017attention}.
Our tokenizer is the same as that of Qwen3.

We do not use grouped-query attention in our experiments, as it introduces an additional hyperparameter (the number of query groups) that complicates the analysis.
We also do not consider shared experts \cite{dai2024deepseekmoe} in this work.

The parameters we vary are listed below:
\begin{itemize}
    \item Number of layers $l$
    \item Hidden dimension $d$
    \item MoE hidden dimension $d_{exp}$ or granularity $g$
    \item Number of experts $n_{exp}$
    \item Number of activated experts $n_{topk}$
\end{itemize}

We fix the attention head number to 4 for all models \cite{porian2024resolving}.
The main model configurations used in Section \ref{sec:main_results} is summarized in Table \ref{tab:model_config} ($g$ fixed to 4).

\begin{table}[h]
\centering
\begin{tabular}{lll}
\hline
$l$ & $d$ & $N_{total}$ for (128,8)  \\ \hline
6 & 288 & 49,766,400  \\
6 & 384 & 88,473,600  \\
8 & 384 & 117,964,800 \\
8 & 512 & 209,715,200  \\
10 & 640 & 409,600,000  \\
14 & 768 & 825,753,600  \\
16 & 1024 & 1,677,721,600 \\
\hline
\end{tabular}
\caption{Model configuration used in experiments in Section \ref{sec:main_results}.$N_{total}$ is calculated with $g=4, n_{exp}=128,n_{topk}=8$.}
\label{tab:model_config}
\end{table}
\subsection{Experimental Setup}

Here, we give a more detailed description of the experimental setup.
Our experiments are conducted using the Megatron-LM framework \cite{shoeybi2019megatron}.
Models are trained with mixed precision.
We also use optimization libraries like FlashAttention \cite{dao2022flashattention} and TransformerEngine \footnote{\url{https://github.com/NVIDIA/TransformerEngine}}.

We adopt the WSD learning rate schedule for all experiments.
In the final phase of training, the learning rate decays linearly to 10\% of its peak value, with the decay phase spanning roughly 10\% of the total training steps, following the setup in \citet{hagele2024scaling}.
To emulate varying token budgets, we save intermediate checkpoints and perform continued learning on them with decayed learning rates.
We set warmup iterations, batch size and learning rate as 100, 2048 and $10^{-3}$ respectively for most of the experiments unless stated otherwise.
This leads to more stable fitting \cite{li2025mis}.
Additional configurations are as in Table \ref{tab:train_config}.

\begin{table*}[h]
\centering
\begin{tabular}{ll}
\hline
\textbf{Configuration} & \textbf{Value} \\
\hline
Context length & 2048 \\
Embedding & Tied \\
Optimizer & AdamW \cite{loshchilov2017fixing} \\
Adam $\beta_1$ & 0.9 \\
Adam $\beta_2$ & 0.95 \\
Weight decay & 0.1 \\
Gradient clipping & 1.0 \\
MoE load balancing loss coefficient & $10^{-3}$ \\
\hline
\end{tabular}
\caption{Training configurations.}
\label{tab:train_config}
\end{table*}

\section{More on Ablation Studies}
\label{app:ablation_details}

\paragraph{Granularity.}
Table~\ref{tab:detail_ablate_g} additionally shows the loss difference in percentage compared to the best configuration in Table~\ref{tab:ablate_g}.
We further note that the uncertainty in loss values due to random seeds is around 0.4\%. 
The smaller model ($l=8,d=384$) is trained with 9B tokens.
For the larger model ($l=18,d=1024$), we increase the training tokens and warmup steps to 46B and 500 respectively.

\paragraph{Width-to-depth ratio.}
Table~\ref{tab:detail_widthdepthratio} shows more results of the width-to-depth ratio ablation study and in more detail, where we additionally show the percentage difference in total and active parameters compared to the best configuration in Table~\ref{tab:ablate_widthdepthratio} (because we cannot obtain the exact values while varying model architecture).
We further note that the uncertainty in loss values due to random seeds is around 0.8\%. 
Models are trained for 4B tokens.

\begin{table*}
\centering
\begin{tabular}{ccccccc}
\hline
$l$ & $d$ & $g\;(d/d_{moe})$ & $n_{exp}$ & $n_{topk}$ & Loss & Loss diff (\%) \\ \hline
8 & 384 & 1 & 32 & 2 & 2.995& 1.7\% \\
8 & 384 & 2 & 64 & 4 & 2.966& 0.7\% \\
8 & 384 & 4 & 128 & 8 & 2.960& 0.48\% \\
8 & 384 & 8 & 256 & 16 & 2.945& 0.00\% \\
8 & 384 & 16 & 512 & 32 & 2.951& 0.21\% \\ \hline
18 & 1024 & 1 & 32 & 2 & 2.334& 1.64\% \\
18 & 1024 & 2 & 64 & 4 & 2.320& 1.04\% \\
18 & 1024 & 4 & 128 & 8 & 2.297& 0.00\% \\
18 & 1024 & 8 & 256 & 16 & 2.304& 0.29\% \\
18 & 1024 & 16 & 512 & 32 & 2.310& 0.58\% \\ \hline
\end{tabular}
\caption{Detailed results for the granularity $g$ ablation study with fixed total and active parameters.}
\label{tab:detail_ablate_g}
\end{table*}

\begin{table*}
 \begin{tabular}{rrrrrllll}
    \hline
$l$ & $d$ & $\gamma(d/l)$ & $n_{nexp}$ & $n_{topk}$ & Loss & Loss diff (\%) & $N_{active}$ diff (\%) & $N_{total}$ diff (\%) \\
    \hline
16 & 272 & 17  & 32 & 2 & 3.19 & 0.19\% & 2.98\% & 1.24\% \\
8  & 384 & 48  & 32 & 2 & 3.21 & 0.84\% & 2.62\% & 0.89\% \\
4  & 544 & 136 & 32 & 2 & 3.20 & 0.50\% & 2.98\% & 1.24\% \\
16 & 240 & 15  & 43 & 4 & 3.21 & 0.65\% & 2.04\% & 2.04\% \\
8  & 336 & 42  & 43 & 4 & 3.19 & 0.00\% & 0.00\% & 0.00\% \\
4  & 480 & 120 & 43 & 4 & 3.22 & 1.12\% & 2.04\% & 2.04\% \\
16 & 160 & 10  & 103 & 16 & 3.27 & 2.68\% & 3.66\% & 1.65\% \\
8  & 224 & 28  & 103 & 16 & 3.26 & 2.20\% & 1.59\% & -0.38\% \\
4  & 320 & 80  & 103 & 16 & 3.24 & 1.68\% & 3.66\% & 1.65\% \\
    \hline
\end{tabular}
\caption{Detailed results for the width-to-depth ratio $d/l$ ablation study with approximately fixed total and active parameters. $g$ is fixed to 4.}
\label{tab:detail_widthdepthratio}
\end{table*}

\section{More on Scaling Laws}
\label{app:scaling_law_details}
\subsection{MoE Design}
Here, we give more details about the experiments conducted in Figure \ref{fig:loss_vs_ntotal} in Section \ref{sec:main_results}.
The ($n_{exp}, n_{topk}$) configurations used are (32,2), (64,2), (64,4), (128,2), (128,8), (256,2), (256,4), (256,8), and (256,16), combined with the core dimensions in Table \ref{tab:model_config}.
All models are trained for 9B tokens.

For fitting the scaling laws, we take the logarithm of all variables and perform linear regression using the ordinary least squares (OLS) method.
We use the \texttt{statsmodels} library for fitting and statistical tests \cite{seabold2010statsmodels}.
The test of significance for each variable is performed using t-tests, with a significance level of 0.05.

We show more combinations of variables tried in Table \ref{tab:fitting_results} in Table \ref{tab:more_fitting_results}, particularly repeating the same combinations from Table \ref{tab:fitting_results} but including more combinations involving $N_{active}$.
We see that some combinations, e.g., $(N_{total}, N_{active},n_{exp})$ and $(N_{total}, N_{active},n_{topk})$ are also quite effective (although slightly smaller $R^2$).
However, our choice of $(N_{total}, n_{exp}, n_{topk})$ is more interpretable, as it is connected to sparsity $s$ directly (of which the component variables are easy to configure), additionally allowing the mathematical interpretation of the $n_{exp}$ penalty as described in the main text.

\begin{table*}[ht]
\centering
\begin{tabular}{@{}lll@{}}
\hline
\textbf{Variables} & \textbf{$R^2$} & \textbf{Result/Intrepretation} \\ \hline
\textit{Simple Baseline} & & \\
$\log(N_{total})$ only & 0.926 & Baseline (Total params). \\
$\log(N_{active})$ only & 0.641 & Baseline (Active params). \\ \hline
\textit{Core Functions} & & \\
$\log(N_{active}) + \log(s)$ & 0.944 & Moderately good fit. \\
$\log(N_{total}) + \log(s)$ & 0.983 & Good fit. \\
$\log(N_{total}) + \log(n_{{exp}}) + \log(n_{topk})$ & \textbf{0.985} & \textbf{Good and disambiguated fit.} \\
$\log(N_{active}) + \log(n_{{exp}}) + \log(n_{topk})$ & 0.981 & Moderately good fit. \\ \hline
\textit{Interaction Functions} & & \\
$\log(N_{total}) + \log(s) + \log(N_{total})\log(s)$ & 0.983 & Interaction term redundant. \\
$\log(N_{total}) + \log(N_{active}) + \log(N_{total})\log(N_{active})$ & 0.988 & Strong multicollinearity problem. \\ \hline

\textit{Other Combinations} & & \\
$\log(N_{total}) + \log(N_{active}) + \log(s)$ & 0.985 &  $N_{active}$'s significance is low. \\
$\log(N_{total}) + \log(N_{active})$ & 0.979 & Moderately good fit. \\
$\log(N_{total}) + \log(N_{active})+ \log(n_{{exp}})$ & 0.983 & Good fit. \\
$\log(N_{total}) + \log(N_{active})+ \log(n_{topk})$ & 0.984 & Good fit. \\
\end{tabular}
\caption{Results of fitting various combinations of variables to the loss values.}
  \label{tab:more_fitting_results}
\end{table*}

\begin{figure}
  \includegraphics[width=\columnwidth]{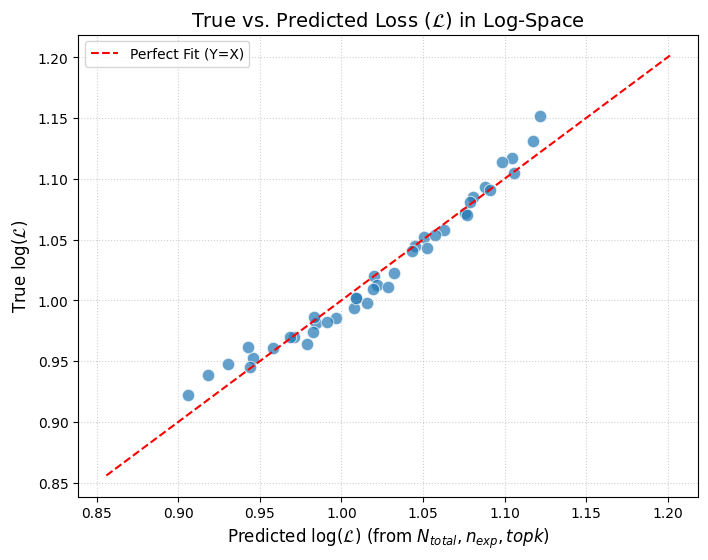}
  \caption{Goodness of fit for the selected scaling law in Section \ref{sec:main_results}.}
  \label{fig:valid_design}
\end{figure}

We further visualize the goodness of fit for the selected scaling law in Figure \ref{fig:valid_design}.

\subsection{Derivation of the $n_{exp}$ Penalty under Fixed Constraints}
\label{app:math}
To understand why increasing the number of experts $n_{exp}$ penalizes performance when total parameters $N_{total}$ and expert sparsity $s$ are fixed, we examine the relationship between active parameters $N_{active}$ and $n_{exp}$.

We start from parameter countings, reiterating the equations for total and active parameters:
$$N_{total} \approx ld^2\left(4 + \frac{3n_{exp}}{g}\right)$$
$$N_{active} \approx ld^2\left(4 + \frac{3n_{topk}}{g}\right)$$
Given that expert sparsity is defined as $s = n_{exp}/n_{topk}$, we substitute $n_{topk} = n_{exp}/s$.
 To isolate the effect of $n_{exp}$ while holding $N_{total}$ constant, we first solve for the core dense term $ld^2$:
 $$ld^2 \approx \frac{N_{total}}{4 + \frac{3n_{exp}}{g}}$$Substituting this back into the equation for $N_{active}$:
 $$N_{active} \approx \left( \frac{N_{total}}{4 + \frac{3n_{exp}}{g}} \right) \cdot \left( 4 + \frac{3n_{exp}}{gs} \right)$$
 The ratio of active to total parameters is thus:
 $$\frac{N_{active}}{N_{total}} \approx \frac{4 + \frac{3n_{exp}}{gs}}{4 + \frac{3n_{exp}}{g}}$$
 In the standard MoE regime where $s > 1$, the denominator grows faster than the numerator as $n_{exp}$ increases. 
 Consequently, $N_{active}$ must decrease to satisfy the constant $N_{total}$ constraint. This confirms that for a fixed memory budget, a larger total number of experts mathematically necessitates a reduction in the model's active compute, leading to the observed performance penalty.

\subsection{Architectural Optimization Routine}
\label{app:opt}
To determine the optimal configuration $(l, d, n_{exp}, n_{topk})$, we treat the expert count $n_{exp}$ and the width-to-depth ratio $\gamma = d/l$ as hyperparameters.
 For each combination, we solve for the maximum dimensions allowed by the memory and inference constraints, denoted by $C_{total}$ and $C_{active}$.
The pseudocode is given in Algorithm \ref{alg:opt}.
In plain language, for each $n_{exp}$ and $\gamma$, we first solve for the maximum $l$ satisfying the total capacity constraint.
 Then, we solve for the maximum $d$ satisfying the same constraint while ensuring that $d$ is aligned to $k_{align}$ (to account for attention head partitioning or hardware alignment requirements).
 Finally, we solve for the maximum $n_{topk}$ satisfying the inference constraint.

 The scaling law $L \propto N_{total}^{-0.052} n_{exp}^{0.023} n_{topk}^{-0.018}$ reveals a critical trade-off. 
 While increasing $n_{exp}$ allows for a larger $N_{total}$ within a fixed memory budget, the positive exponent ($0.023$) acts as a penalty.
  The optimization loop ensures that the gain from $N_{total}$ outweighs this penalty, while $n_{topk}$ is used as the final lever to recover performance within the inference constraint.

  \begin{algorithm}
\caption{MoE Architectural Optimization}
\label{alg:opt}
\begin{algorithmic}[1]
\STATE \textbf{Input:} Memory constraint $C_{total}$, Inference constraint $C_{active}$, Alignment factor $k_{align}$
\STATE \textbf{Parameters:} $\Gamma \in [32, 64]$, $n_{exp} \in \{2^1, 2^2, \dots, 2^k\}$
\STATE \textbf{Initialize:} $L_{min} \leftarrow \infty, \theta^* \leftarrow \emptyset$
\FOR{each $n_{exp} \in \{2^1, \dots, 2^k\}$}
    \FOR{each $\gamma \in \Gamma$}
        \STATE $l \leftarrow \lfloor (C_{total} / (\gamma^2 (4 + 0.75 n_{exp})))^{1/3} \rfloor$
        \STATE $d \leftarrow \text{round}(\gamma \cdot l / k_{align}) \cdot k_{align}$
        \WHILE{$l \cdot d^2 \cdot (4 + 0.75 n_{exp}) > C_{total}$}
            \STATE $d \leftarrow d - k_{align}$
        \ENDWHILE
        \STATE $n_{topk} \leftarrow \min \left( n_{exp}, \lfloor \frac{4}{3} (\frac{C_{active}}{l d^2} - 4) \rfloor \right)$
        \IF{$n_{topk} \ge 1$}
            \STATE $L \leftarrow (l d^2 (4 + 0.75 n_{exp}))^{-0.052} \cdot n_{exp}^{0.023} \cdot n_{topk}^{-0.018}$
            \IF{$L < L_{min}$}
                \STATE $L_{min} \leftarrow L, \theta^* \leftarrow (l, d, n_{exp}, n_{topk})$
            \ENDIF
        \ENDIF
    \ENDFOR
\ENDFOR
\RETURN $\theta^*$
\end{algorithmic}
\end{algorithm}

\paragraph{Comparison with Qwen3-235B-A22B.}

Using the optimization routine, we derive an MoE configuration under similar memory and inference constraints as Qwen3-235B-A22B.
  Using the Qwen3 memory and inference constraints, our optimization routine suggests an MoE configuration of $(l,d,n_{exp},n_{topk}) = (94, 4096, 128, 7)$, resulting in 234B total parameters and 21.7B active parameters.
  This configuration is close to that of Qwen3-235B-A22B but with more layers, which uses $(l,d,n_{exp},n_{topk}) = (83, 5312, 128, 8)$.
  We also note that Qwen3-235B-A22B uses $g=2.7$.

\subsection{Testing the MoE Design Principle}
\label{app:test_moe}
We describe the settings for the experiments in Figure \ref{fig:128_8_vs_256_16} in Section \ref{sec:main_results}.
The models (configurations in Table \ref{tab:model_config}) are trained with various dataset sizes (from 9B to 50B tokens).

\begin{figure}
  \includegraphics[width=\columnwidth]{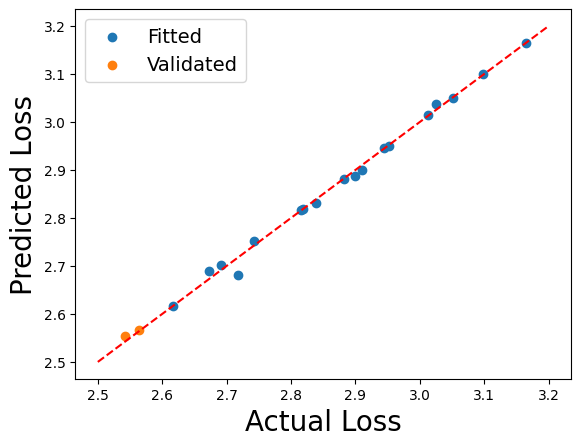}
  \caption{Goodness of fit for the (128,8) configuration.}
  \label{fig:fit_128}
\end{figure}

\begin{figure}
  \includegraphics[width=\columnwidth]{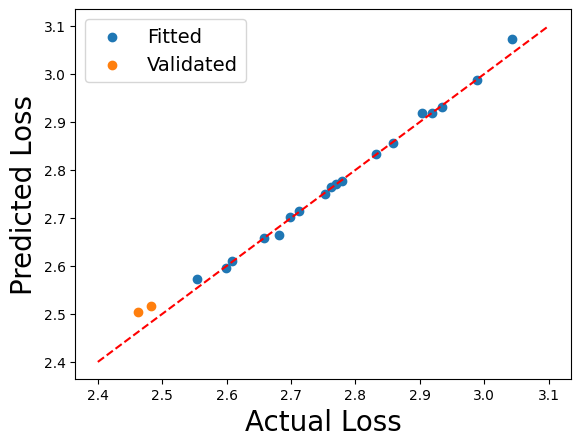}
  \caption{Goodness of fit for the (256,16) configuration.}
  \label{fig:fit_256}
\end{figure}

We fit the losses with the functional form of Chinchilla-style scaling laws following \cite{hoffmann2022training,besiroglu2024chinchilla}: We perform optimization using the Huber loss ($\delta=10^{-3}$) and the BFGS algorithm, to fit the logarithm of the loss via the LogSumExp trick applied to the RHS of functional forms.
The fitted coefficients are as follows:
\begin{itemize}
    \item For the (128,8) configuration: $A=28, B=229, E=1.08, \alpha=0.28, \beta=0.16$
    \item For the (256,16) configuration: $A=564, B=640,500, E=2.0, \alpha=0.64, \beta=0.36$
\end{itemize}
We further visualize the goodness of fit in Figure \ref{fig:fit_128} and \ref{fig:fit_256}.


\end{document}